\documentclass[aps,pre,showpacs,groupedaddress]{revtex4}
\usepackage{amsfonts}
\usepackage{mathrsfs}
\usepackage{graphicx}
\usepackage{amsmath}
\usepackage{amssymb}
\usepackage{amsbsy}
\usepackage{bm}

\def\bs{\mathbf{s}}
\def\bw{\mathbf{w}}
\def\bh{\mathbf{h}}
\def\bsg{\boldsymbol{\sigma}}

\begin{document}

\title{Mechanisms of dimensionality reduction and decorrelation in deep neural networks}

\author{Haiping Huang}
\email{physhuang@gmail.com;www.labxing.com/hphuang2018}
\affiliation{School of Physics,
Sun Yat-sen University, Guangzhou 510275, People's Republic of China}
\affiliation{Laboratory for Neural Computation and Adaptation, RIKEN Center for Brain Science, Wako-shi, Saitama
351-0198, Japan}

\date{\today}

\begin{abstract}
Deep neural networks are widely used in various domains. However, the nature of computations at each layer of the deep networks is far from being well
understood. Increasing the interpretability of deep neural networks is thus important. Here, we construct a mean-field framework to understand how compact representations
are developed across layers, not only in deterministic deep networks with random weights but also in generative deep networks where an unsupervised learning is carried out.
Our theory shows that the deep computation implements a dimensionality reduction while maintaining a finite level of weak correlations between neurons for possible feature extraction. 
Mechanisms of dimensionality reduction and decorrelation are unified in the same framework.
This work may pave the way for understanding how a sensory hierarchy works.
\end{abstract}

 \maketitle

\textit{Introduction.}---The sensory cortex in the brain encodes the structure of the environment in an efficient way. This is achieved by creating progressively better representations of 
sensory inputs, and these representations finally become easily decoded without any reward or supervision signals~\cite{DiCarlo-2007,DiCarlo-2012,Krieg-2013}. This kind of learning is called unsupervised learning, which
has long been thought of as a fundamental function of the sensory cortex~\cite{Marr-1970}.
Based on the similar computational principle, many layers of artificial neural networks were designed to perform a non-linear dimensionality reduction of high dimensional data~\cite{Hinton-2006a},
which later triggered resurgence of deep neural networks. By stacking unsupervised modules on top of each other, one can produce a deep feature hierarchy, in which high-level features can be
constructed from less abstract ones along the hierarchy. However, these empirical results do not have a principled understanding so far. Understanding what each layer exactly computes may shed light on how sensory systems work in general.

Recent theoretical efforts focused on the layer-wise propagation of one input vector length, correlations between two inputs~\cite{NIPS-2016}, and clustered noisy inputs of supervised classification tasks~\cite{Somp-2014,Somp-2016}, generalizing a theoretical work of
layered feedforward neural networks that studied the iteration of the overlap between layer's activity and embedded random patterns~\cite{JPA-1989}.
However, these studies did not address covariance of neural activity, one
important feature of neural data modeling~\cite{Cunning-2017}, which is directly related to the dimensionality and complexity of hierarchical representations. 
Therefore, a clear understanding of hierarchical representations has been lacking so far, which makes deep computation extremely non-transparent.
Here, we propose a mean-field theory of input dimensionality reduction in deep neural networks. In this theory, we capture how a deep non-linear transformation reduces
the dimensionality of a data representation, and moreover, how the covariance level (redundancy) varies along the hierarchy. Both of these two features are fundamental
properties of deep neural networks, and even information processing in vision~\cite{grad-2015}.

Our theory helps to advance the understanding of deep computation in two aspects: (i) There exists an operating point where input and output
covariance levels are equal. This point controls the level of covariance neither diverging nor decaying to zero, given sufficiently strong connections between layers. (ii) The dimensionality
of data representation is reduced across layers, due to an additive positive term (contributed by the previous layer) affecting the dimensionality in a divisive way.
These computational principles are revealed not only in deterministic deep networks with random weights but also in generative deep trained networks.
Our analytical findings coincide with numerical simulations, demonstrating that the previous empirically observed dimensionality reduction~\cite{Hinton-2006a,DiCarlo-2007,DiCarlo-2012} and the redundancy reduction hypothesis~\cite{Barlow-1961} could be
theoretically explained within the same mean-field framework.

\begin{figure}
    \includegraphics[bb=87 357 420 685,scale=0.4]{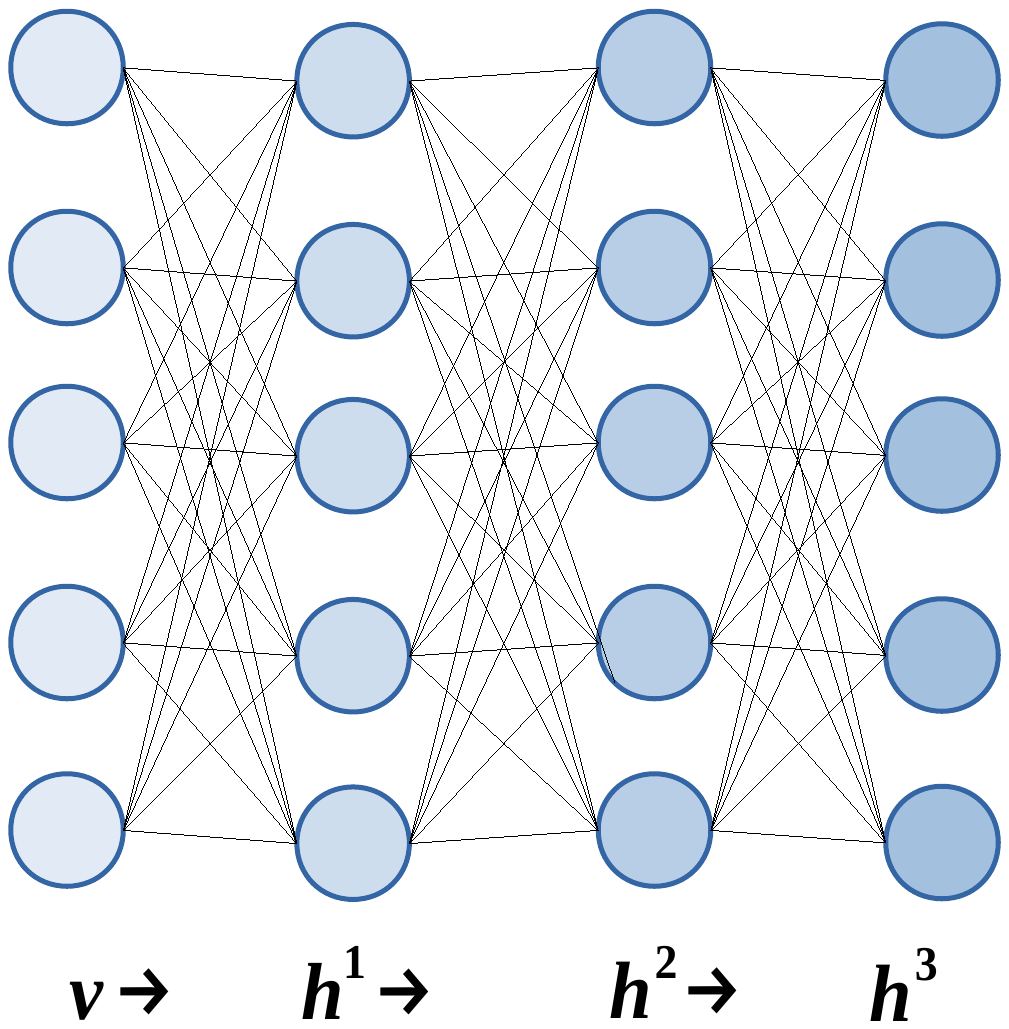}
  \caption{ (Color online) Schematic illustration of a deep neural network. The deep neural network performs a layer-by-layer non-linear transformation of the original
  input data (a high dimensional vector $\mathbf{v}$). During the transformation, a cascade of internal representations ($\mathbf{h}^1,\cdots,\mathbf{h}^d$) are created. Here, $d=3$ denotes
  the depth of the deep network.
  }\label{dnn}
\end{figure}

\textit{A deterministic deep network.}---A deep network is a multi-layered neural network performing hierarchical non-linear transformations of sensory inputs (Fig.~\ref{dnn}).
The number of hidden layers is defined as the depth of the network, and the number of neurons at each layer is called the width of that layer. For simplicity, we assume an equal width ($N$).
Weights between $l-1$ and $l$-th layers are specified by a matrix $\mathbf{w}^l$, 
in which the $i$-th row corresponds to incoming connections to the neuron $i$ at the higher layer.
Biases of neurons at the $l$-th
layer are denoted by $\mathbf{b}^l$. The input data vector is denoted by $\mathbf{v}$, and 
$\mathbf{h}^l$ ($l=1,\cdots,d$) denotes a hidden representation of the $l$-th layer, in which each entry $h_i^l$ defines a non-linear transformation of its pre-activation $\tilde{a}_i^l\equiv[\mathbf{w}^l\mathbf{h}^{l-1}]_i$, as
$h_i^l=\phi(\tilde{a}_i^l+b_i^l)$. Without loss of generality, we choose the non-linear transfer
function as $\phi(x)=\tanh(x)$, and assume that the weight follows a normal distribution $\mathcal{N}(0,g/N)$, and the bias follows $\mathcal{N}(0,\sigma_b)$. Random weight assumption plays an important role in recent 
studies of artificial neural networks~\cite{Monasson-2017,Sohl-2017,Barra-2018,Huang-2015b,Huang-2016b,Huang-2017}, and the weight distribution of trained networks may appear random~\cite{Random-2017}.

We consider a Gaussian input ensemble with zero mean, covariance $\left<v_iv_j\right>=\frac{r_{ij}}{\sqrt{N}}$ for all $i\neq j$ ($r_{ij}$ is a uniformly-distributed random variable from $[-\rho,\rho]$), and variance $\left<v_i^2\right>=1$.
In the following derivations, we define the weighted-sum $\tilde{a}_i^l$ subtracted by its mean as $a_i^l=\sum_jw^l_{ij}(h^{l-1}_j-\left<h_j^{l-1}\right>)$, thus $a_i^l$ has zero mean.
As a result, the covariance of $\mathbf{a}^l$ can be expressed as $\Delta^l_{ij}=\left<a_i^la_j^l\right>=\left[\mathbf{w}^l\mathbf{C}^{l-1}(\mathbf{w}^l)^T\right]_{ij}$, where $\mathbf{C}^{l-1}$
defines the covariance matrix of neural activity at $l-1$-th layer (also called connected correlation matrix in physics). Because the deep network defined in Fig.~\ref{dnn} is a fully-connected
feedforward network, where each neuron at an intermediate layer receives a large number of inputs, the central limit theorem implies that the mean of hidden neural activity 
$\mathbf{m}^l$ and covariance $\mathbf{C}^l$ are given separately by
\begin{subequations}\label{dMFT}
\begin{align}
 m_i^l&=\left<h_i^l\right>=\int Dt\phi\left(\sqrt{\Delta^l_{ii}}t+[\mathbf{w}^l\mathbf{m}^{l-1}]_i+b_i^l\right),\\
 \begin{split}
 C^l_{ij}&=\int DxDy\phi\left(\sqrt{\Delta_{ii}^l}x+b_i^l+[\mathbf{w}^l\mathbf{m}^{l-1}]_i\right)\\
 \times&\phi\left(\sqrt{\Delta_{jj}^l}(\Psi x+y\sqrt{1-\Psi^2})+b_j^l+[\mathbf{w}^l\mathbf{m}^{l-1}]_j\right)-m_i^lm_j^l,
 \end{split}
 \end{align}
\end{subequations}
where $Dx=e^{-x^2/2}dx/\sqrt{2\pi}$, and $\Psi=\frac{\Delta_{ij}^l}{\sqrt{\Delta_{ii}^l\Delta_{jj}^l}}$. To derive Eq.~(\ref{dMFT}), we parametrize $a_i^l$ and $a_j^l$ by independent normal random variables (see appendix~\ref{MFE}). Eq.~(\ref{dMFT}) forms 
an iterative mean-field equation across layers to
describe the transformation of the activity statistics in deep networks. 

To characterize the collective property of the entire hidden representation, we define an intrinsic dimensionality of the representation as
 $D=\frac{\left(\sum_{i=1}^N\lambda_i\right)^2}{\sum_{i=1}^N\lambda_i^2}$~\cite{NIPS2010},
where $\{\lambda_i\}$ is the eigen-spectrum of the covariance matrix $\mathbf{C}^l$. It is expected that $D=N$ if each component of the representation is generated
independently with the same variance. Generally speaking, non-trivial correlations in the representation will result in $D<N$. Therefore, we can use
the above mean-field equation together with the dimensionality to address how the complexity of hierarchical representations changes along the depth.

\begin{figure}
     \includegraphics[bb=0 0 419 316,scale=0.55]{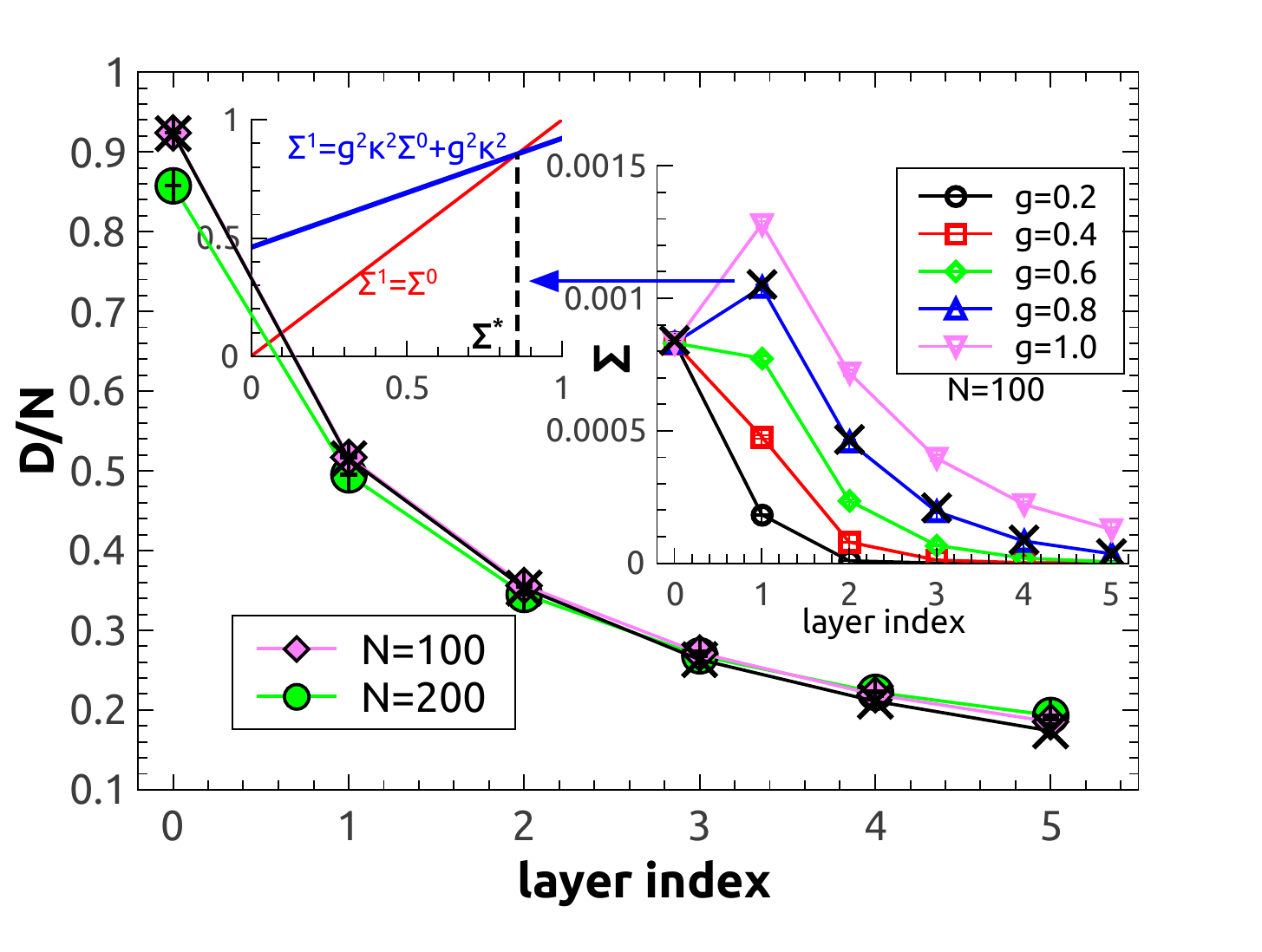}
  \caption{
  (Color online) Representation dimensionality versus depth in deterministic deep networks with random weights. Ten network realizations are considered for each network width. $\rho/\sqrt{N}=0.05$, $g=0.8$, and
  $\sigma_b=0.1$. The right inset shows how the overall strength of covariance
  changes with depth and connection strength ($g$), and the left inset is a mechanism illustration ($\Sigma^{0,1,*}$ has been scaled by $N$). The crosses show simulation results ($g=0.8$) obtained from $10^5$ sampled configurations at each layer,
  compared with the theoretical predictions.
  }\label{ddnres}
\end{figure}

Based on this mean-field framework, we first study the aforementioned deterministic deep neural networks. Regardless of whichever network width used, we find that the representation dimensionality progressively decreases across layers (Fig.~\ref{ddnres}). 
The theoretical results agree very well with numerical simulations (indicated by crosses in Fig.~\ref{ddnres}).
This shows that, even in a random multi-layered neural network, a more compact representation of 
the correlated input is gradually computed as the network becomes deeper, which is also one of basic properties in biological hierarchical computations~\cite{DiCarlo-2012,Yamin-2016}.

To get deeper insights about the hidden representation, we study how the overall strength of covariance at each layer changes
with the network depth and connection strength ($g$). The overall covariance-strength is
measured by $\Sigma=\frac{2}{N(N-1)}\sum_{i<j}C_{ij}^2$, which is related to the dimensionality via $\Sigma=\frac{1}{N-1}\Bigl[(\frac{1}{N}\sum_iC_{ii})^2/\tilde{D}-\frac{1}{N}\sum_iC_{ii}^2\Bigr]$ where $\tilde{D}=D/N$, which is derived by noting that ${\rm tr}(\mathbf{C})=\sum_i\lambda_i$ and ${\rm tr}(\mathbf{C}^2)=\sum_i\lambda_i^2$. 
We find that, to support an effective representation where neurons are
not completely independent, the connection strength must be sufficiently strong (Fig.~\ref{ddnres}), such that weakly-correlated neural activities are
still maintained at later stages of processing. Otherwise, the information will be blocked from passing through that layer where
the neural activity becomes completely independent. High correlations imply strong statistical dependence, and thus
redundancy. An efficient representation must not be highly redundant~\cite{Barlow-1961}, because a highly redundant representation can not be
easily disentangled and is thus not useful for computation, e.g., co-adaptation of neural activities is harmful for feature extraction~\cite{dropout}.

The dimensionality reduction results from the nested non-linear transformation of input data.
For a mechanistic explanation, by noting that $\Delta_{ij}^l$ is of the order $O(1/\sqrt{N})$, we expand $C_{ij}^l=K_{ij}\Delta_{ij}^l+O((\Delta_{ij}^l)^2)$ in a large-$N$ limit (appendix~\ref{Limit}), where $K_{ij}\equiv\phi'(x_i^0)\phi'(x_j^0)$, and $x_{i,j}^0\equiv b_{i,j}^l+[\mathbf{w}^l\mathbf{m}^{l-1}]_{i,j}$.
Then $\Sigma^l\simeq g^2\kappa^2\Sigma^{l-1}+\frac{g^2\kappa^2}{N^2}\sum_i(C_{ii}^{l-1})^2$ where $\kappa\equiv\overline{(\phi'(x_i^0))^2}$ (the average is taken over the random network parameters, see appendix~\ref{Limit}).
For the random model, $N\Sigma^1=g^2\kappa^2(N\Sigma^0+1)$, which determines a critical $N\Sigma^*=\frac{g^2\kappa^2}{1-g^2\kappa^2}$ (so-called operating point), such that a first boost of the correlation strength is observed 
when $\Sigma^0<\Sigma^*$ (Fig.~\ref{ddnres}); otherwise, the correlation level is maintained, or decorrelation is achieved. The iteration of $\Sigma^l$ can be used to derive $\tilde{D}^1=\frac{1}{(N-1)\Sigma^0+1+\Upsilon}$ where $\Upsilon>0$ and its value depends on the layer's parameters.
Thus $\Upsilon$ determines how significantly the dimensionality is reduced, 
and the dimensionality reduction is explained as $\tilde{D}^1<\tilde{D}^0=\frac{1}{(N-1)\Sigma^0+1}$.
This relationship carries over to deeper layers (appendix~\ref{Limit}), due to an additive positive term in the denominator of the dimension formula. Therefore, the decorrelation of redundant inputs together with the dimensionality reduction is theoretically explained.

\textit{A stochastic deep network.}---It is of practical interest to see
whether a deep generative model trained in an unsupervised way has the similar collective behavior. We consider a deep belief network (DBN) as a typical example of 
stochastic deep networks~\cite{Hinton-2006a}, in which each neuron's activity at one hidden layer takes a binary value ($\pm1$) according to
a stochastic function of the neuron's pre-activation.
Specifically, the DBN is composed of multiple restricted Boltzmann machines (RBMs) stacked on top of each other (Fig.~\ref{dnn}). 
RBM is a two-layered neural network, where there are no lateral connections within each layer, and the bottom (top) layer is also named the visible (hidden) layer. Therefore,
given the input $\mathbf{h}^l$ at $l$-th layer, the neural representation
at a higher ($l+1$-th) layer is determined by a conditional probability
\begin{equation}\label{dbn}
 P(\bh^{l+1}|\bh^l)=\prod_i\frac{e^{h_i^{l+1}([\bw^{l+1}\bh^l]_i+b_i^{l+1})}}{2\cosh([\bw^{l+1}\bh^l]_i+b_i^{l+1})}.
\end{equation}
Similarly, $P(\bh^l|\bh^{l+1})$ is also factorized. 

The DBN as a generative model, once network parameters (weights and biases)
are learned (so-called training) from a data distribution, can be used to reproduce the samples mimicking that data distribution. With deep layers, the network becomes more expressive to capture 
high-order interdependence among components of a high-dimensional input, compared with a shallow RBM network. To study the expressive property of the DBN, we first
specify a data distribution generated by a random RBM whose parameters follow the normal distribution $\mathcal{N}(0,g/N)$ for
weights and $\mathcal{N}(0,\sigma_b)$ for biases. Using the random RBM as a data generator allows us to
calculate analytically the complexity of the input data.
In the random RBM, the hidden neural activity $\bh$ at the top layer can be marginalized over using the conditional independence (Eq.~(\ref{dbn})), thus the distribution of the representation $\mathbf{v}$ at the bottom layer can 
be expressed as (appendix~\ref{train})
\begin{equation}\label{rbmeq}
 P(\mathbf{v})=\frac{1}{Z}\prod_a\left[2\cosh([\bw^{l+1}\mathbf{v}]_a+b_a)\right]\prod_ie^{v_ib_i},
\end{equation}
where $a$ is the site index of hidden neurons, and $Z$ is the partition function. Based on the Bethe approximation~\cite{Bethe}, which captures weak correlations among neurons,
covariance of neural activity (the same definition as before) under Eq.~(\ref{rbmeq}) can be computed from the approximate free energy using the linear response theory (appendix~\ref{CMrbm}).
The estimated statistics of the random-RBM representation are used as a starting point from which the mean-field complexity-propagation equation (Eq.~(\ref{dMFT})) iterates,
for the investigation of the dimensionality and redundancy reduction in
the deep generative model.

\begin{figure}
  \includegraphics[bb=0 0 380 287,scale=0.58]{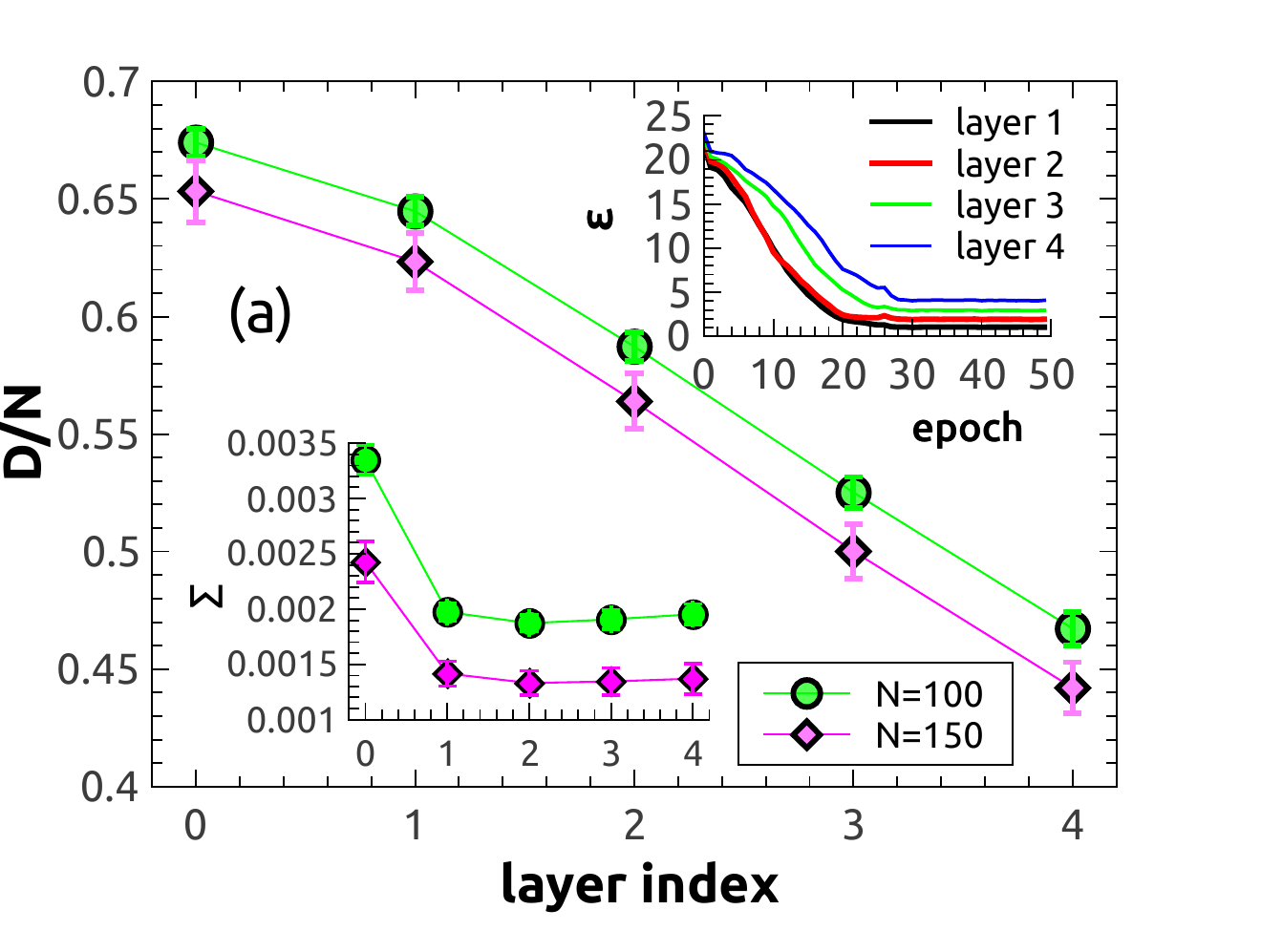}
  \includegraphics[bb=0 0 325 244,scale=0.65]{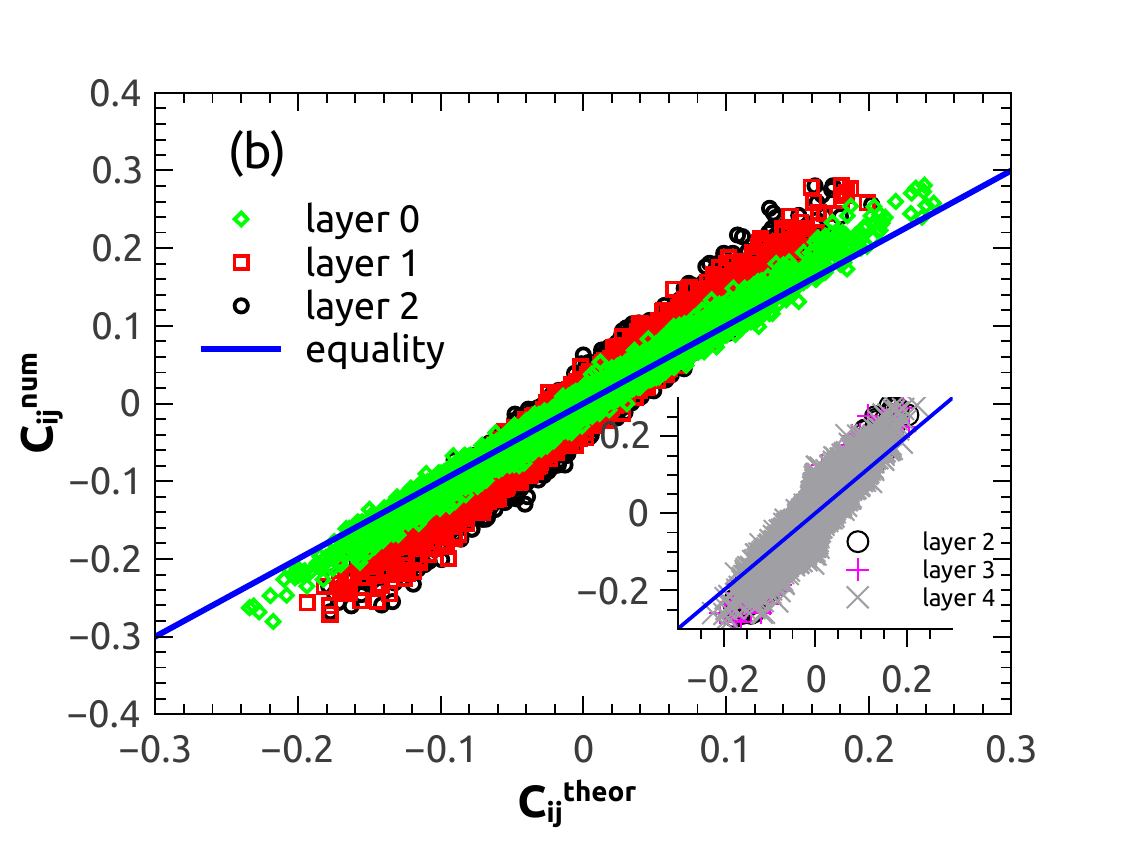}
  \includegraphics[bb=0 0 320 244,scale=0.65]{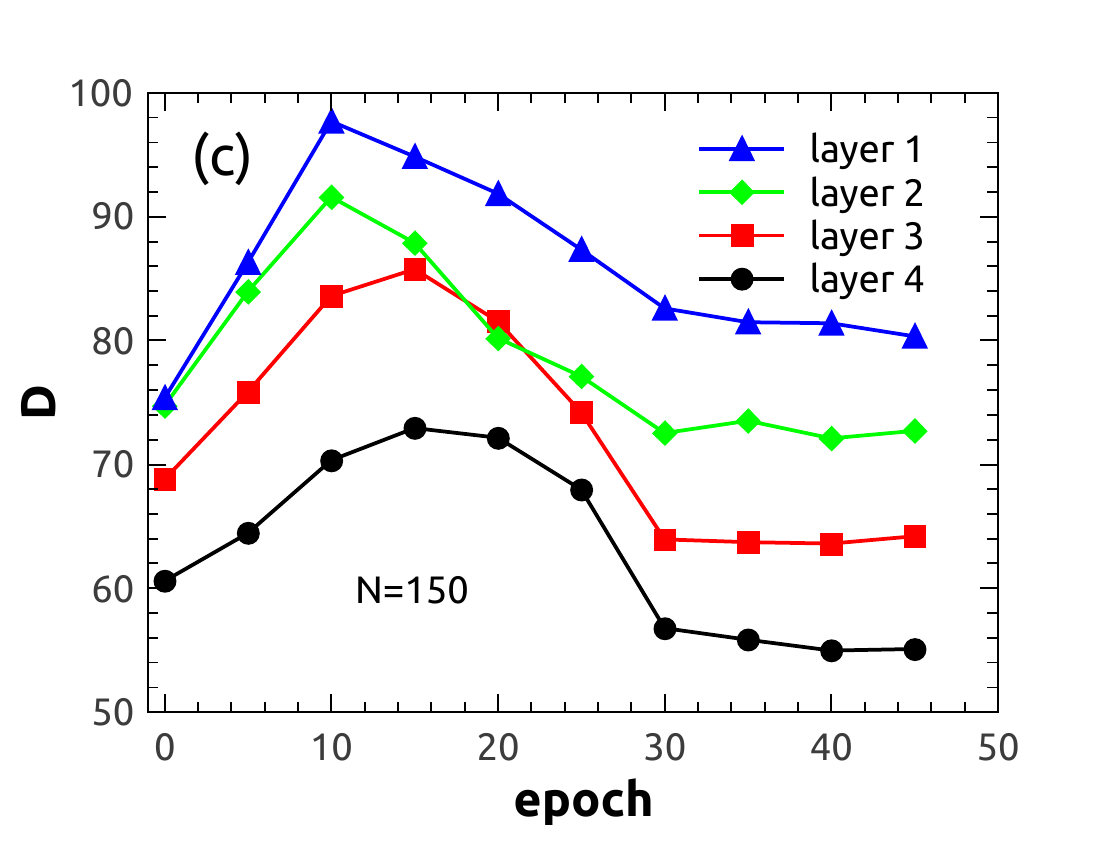}
  \caption{(Color online) (a) Representation behavior as a function of depth in generative deep networks. Ten network realizations are considered for each network width.
 The top inset shows an example ($N=150$) of reconstruction errors ($\varepsilon\equiv\|\bh'-\bh\|^2_2$) between input $\bh$ and reconstructed one $\bh'$ for each layer during learning. The bottom inset shows the overall strength of covariance as
  a function of depth. (b) Numerically estimated off-diagonal correlation versus its theoretical prediction ($N=150$). In these plots, we generate $M=60000$ training examples (each example is an $N$-dimensional vector) from the random RBM whose parameters follow the normal distribution $\mathcal{N}(0,g/N)$ for
weights and $\mathcal{N}(0,\sigma_b)$ for biases. $g=0.8$ and $\sigma_b=0.1$. Then these examples are learned by the DBN
  (see simulation details in appendix~\ref{train}). (c) One typical learning trial shows how the estimated dimensionality evolves.
      }\label{sdnres}
 \end{figure}

Finally, we study the generative deep network where network parameters are learned in a bottom-up pass from the representations at lower layers. The 
network parameters for each stacked RBM in the DBN are updated by contrast divergence procedure truncated to one step~\cite{Hinton-2006b}. With this layer-wise training, each layer
learns a non-linear transformation of the data, and upper layers are conjectured to learn more abstract (complex)
concepts, which is a key step in object recognition problems~\cite{Lee-2011}.
One typical learning trajectory for each layer is shown in the top inset of Fig.~\ref{sdnres} (a), where the reconstruction error decreases with
the learning epoch. The input data and subsequent representation complexity is captured very well by the theory (Fig.~\ref{sdnres} (b)). We use the mean-field framework derived for deterministic networks to
study the complexity propagation (starting from the statistics of the input data), which is reasonable, because to suppress the noise due to sampling, the mean activities at the intermediate layer are used as the input data when the next layer is learned~\cite{Hinton-2006b}.
Therefore, the stochasticity of neural response is implicitly encoded into the learned parameters during training.

Compared to the initial input dimensionality, the representation dimensionality the successive layers create becomes lower (Fig.~\ref{sdnres} (a)), which coincides with observations in 
the deterministic random deep networks. This feature does not change when more neurons are used in each layer. During learning, the evolution of the dimensionality 
displays a non-monotonic behavior (Fig.~\ref{sdnres} (c)): the dimensionality first increases and then decreases to a stationary value. Moreover, the learning decorrelates the correlated input, whereas, after the first drop, the learning seems to preserve
a finite level of correlations (the bottom inset of Fig.~\ref{sdnres} (a)). These compact representations may remove some irrelevant
factors in the input, which facilitates formation of easily-decoded representations at deeper layers. Our theoretical analysis in random neural networks
will likely carry over to this unsupervised learning system, e.g., the operating point may explain the
low-level preserved correlation.

By looking at the eigenvalue distribution of the covariance matrix, we find that the distribution for the unsupervised learning system deviates significantly from
the Marchenko-Pastur law of a Wishart ensemble~\cite{MPlaw} (appendix~\ref{spect}). 
For the random neural networks, the eigenvalue distribution at deep layers seems to assign a higher probability density when the eigenvalue gets close to zero, yet
a lower density at the tail of the distribution, compared with the Marchenko-Pastur law of a random-sample covariance matrix~\cite{Dean-2003} (appendix~\ref{spect}). Therefore, the dimensionality reduction and its relationship with decorrelation
are a nontrivial result of the deep computation.

\textit{Summary.}---Brain computation can be thought of as a transformation of internal representations along different
stages of a hierarchy~\cite{Krieg-2013,Yamin-2016}. Deep artificial neural networks can also be interpreted as a way of creating progressively better 
representations of input sensory data. Our work provides a mean-field evidence about this picture that compact representations of 
relatively low dimensionality are progressively created by deep computation, while a small level of correlations is still maintained to make
feature extraction possible, in accord with the redundancy reduction hypothesis~\cite{Barlow-1961}. In the deep computation, more abstract concepts captured at higher layers along the hierarchy are typically built upon
less abstract ones at lower layers, and high level representations are generally invariant to local changes of the input~\cite{DiCarlo-2012}, which thereby coincides with our theory that demonstrates a compact (compressed) representation formed by a series of
dimensionality reduction. Unwanted variability may be suppressed in this compressed representation. It was hypothesized that
neuronal manifolds at lower layers are strongly entangled with each other, while at later stages, manifolds are flattened to facilitate 
that relevant information can be easily decoded by downstream areas~\cite{DiCarlo-2007,DiCarlo-2012,Yamin-2016}, which connects to the small level of correlations preserved in the network for a representation that may be maximally disentangled~\cite{Soatto-2017}.

Our work thus provides a theoretical underpinning of
the hierarchical representations, through a physics explanation of dimensionality reduction and decorrelation, which encourages several directions such as generalization of this theory to more complex architectures and data distributions, demonstration of how the compact representation helps generalization (invariance) or 
discrimination (selectivity) in a neural system~\cite{DiCarlo-2016,Rust-2013}, and using the revealed principles to control
the complexity of internal representations for an engineering application.

\section*{Acknowledgments}

 I thank Hai-Jun Zhou and Chang-Song Zhou for their insightful comments. This research was supported by AMED under Grant Number JP18dm020700 and the start-up budget 74130-18831109 of
 the 100-talent-program of Sun Yat-sen University.
\appendix

\section{Derivation of mean-field equations for the complexity propagation in deep random neural networks}
\label{MFE}
We first derive the mean-field equation for the mean activity $m_i^l$, by noting that its pre-activation $\tilde{a}_i^l+b_i^l=a_i^l+[\mathbf{w}^l\mathbf{m}^{l-1}]_i+b_i^l$.
$a_i^l$ behaves like a Gaussian random variable with zero mean and variance $\Delta_{ii}^l$, depending on the fluctuating input; thus the average operation for the mean activity can be computed as a Gaussian integral:
\begin{equation}\label{ml}
 m_i^l=\left<h_i^l\right>=\int Dt\phi\left(\sqrt{\Delta^l_{ii}}t+[\mathbf{w}^l\mathbf{m}^{l-1}]_i+b_i^l\right).
\end{equation}
Analogously, to compute the covariance $C_{ij}^l$, one first evaluates the statistics of the pre-activations of unit $i$ and $j$, i.e., $a_i^l$ and $a_j^l$, which follows a joint Gaussian
distribution with zero mean, variance $\Delta_{ii}^l$ and $\Delta_{jj}^l$, respectively, and covariance $\Delta_{ij}^l$. Therefore, one can use two independent standard Gaussian random
variables with zero mean 
and unit variance to parametrize this joint distribution, which results in
\begin{equation}\label{Cl}
 C^l_{ij}=\int DxDy\phi\left(\sqrt{\Delta_{ii}}x+b_i^l+[\mathbf{w}^l\mathbf{m}^{l-1}]_i\right)\phi\left(\sqrt{\Delta_{jj}}\left(\Psi x+y\sqrt{1-\Psi^2}\right)+b_j^l+[\mathbf{w}^l\mathbf{m}^{l-1}]_j\right)-m_i^lm_j^l,
\end{equation}
where $Dx=e^{-x^2/2}dx/\sqrt{2\pi}$, and $\Psi=\frac{\Delta_{ij}}{\sqrt{\Delta_{ii}\Delta_{jj}}}$. The superscript $l$ is omitted for the covariance of $\mathbf{a}^l$.
It is easy to verify that the above parametrization of $a_i^l$ and $a_j^l$ follows the same statistics as mentioned above.

\section{Theoretical analysis in the large-$N$ limit}
\label{Limit}
In the thermodynamic limit, the covariance of activation $\mathbf{a}^l$, $\Delta_{ij}^l$, is of the order of $O(1/\sqrt{N})$, where $N$ is the network width.
This is because both the weights and the (connected) correlations are also of the order of $O(1/\sqrt{N})$. Thus, one can expand the covariance equation (Eq.(\ref{Cl})) in the small
$\Delta_{ij}$ limit. After the expansion and some simple algebra, one obtains
\begin{equation}\label{Cl03}
 C^l_{ij}=K_{ij}\Delta_{ij}+O(\Delta_{ij}^2),
\end{equation}
where $K_{ij}=\phi'(x_i^0)\phi'(x_j^0)$, and $x_{i,j}^0=b_{i,j}^l+[\mathbf{w}^l\mathbf{m}^{l-1}]_{i,j}$. It follows that
\begin{equation}\label{sigmal}
\begin{split}
 \Sigma^l&\simeq\frac{2}{N(N-1)}\sum_{i<j} K_{ij}^2\Delta_{ij}^2\\
 &\simeq g^2\overline{K_{ij}^2}\Sigma^{l-1}+\frac{g^2\overline{K_{ij}^2}}{N^2}\sum_i(C_{ii}^{l-1})^2,
 \end{split}
\end{equation}
where $\overline{K_{ij}^2}\simeq\overline{(\phi'(x_i^0))^2}^2$ in which the mean (overline) is taken over the quenched disorder, based on two facts that the correlation between different weights is negligible, and the covariance of the mean pre-activations of different units is negligible as well in the 
thermodynamic limit.

Finally, $\overline{(\phi'(x_i^0))^2}=\int Dt\int Du(\phi'(\sqrt{\sigma_b}u+\sqrt{gQ^{l-1}}t))^2$, where one independent Gaussian random variable ($u$) corresponds to the randomness of the bias, and the other Gaussian random variable ($t$) corresponds to
the random mean pre-activation $[\mathbf{w}^l\mathbf{m}^{l-1}]_i$ because of the random weights. In addition, $gQ$ specifies the variance of 
the mean pre-activation, with the definition $Q=\frac{1}{N}\sum_im_i^2$ (or spin glass order parameter in physics).
Clearly, $Q$ can be iteratively computed from one layer to its next layer as follows:
\begin{equation}\label{Ql}
 Q^l=\int Dt\int Du\phi^2(\sqrt{\sigma_b}u+\sqrt{gQ^{l-1}}t).
\end{equation}
Note that the initial $Q^0=0$ by the construction of the random model.

Looking at $l=1$ for the random model, one finds $\overline{(\phi'(x_i^0))^2}=\int Du(\phi'(\sqrt{\sigma_b}u))^2\equiv\kappa$ as defined in the main text.
It follows that $\Sigma^1=g^2\kappa^2\Sigma^0+g^2\kappa^2/N$. Following the definition of the normalized dimensionality [$\tilde{D}=\frac{({\rm Tr}(\mathbf{C}))^2}{N{\rm Tr}(\mathbf{C}^2)}$, an alternative definition of the dimensionality in the main text], one easily arrives at
the relationship between $\tilde{D}^1=\frac{1}{(N-1)\Sigma^0+1+\Upsilon}$ and $\tilde{D}^0=\frac{1}{(N-1)\Sigma^0+1}$, noting that
$\Upsilon\equiv\frac{\overline{(\phi'(x_i^0))^4}}{\overline{(\phi'(x_i^0))^2}^2}\geq1$.

\begin{figure}
  \includegraphics[bb=0 0 351 244,scale=0.5]{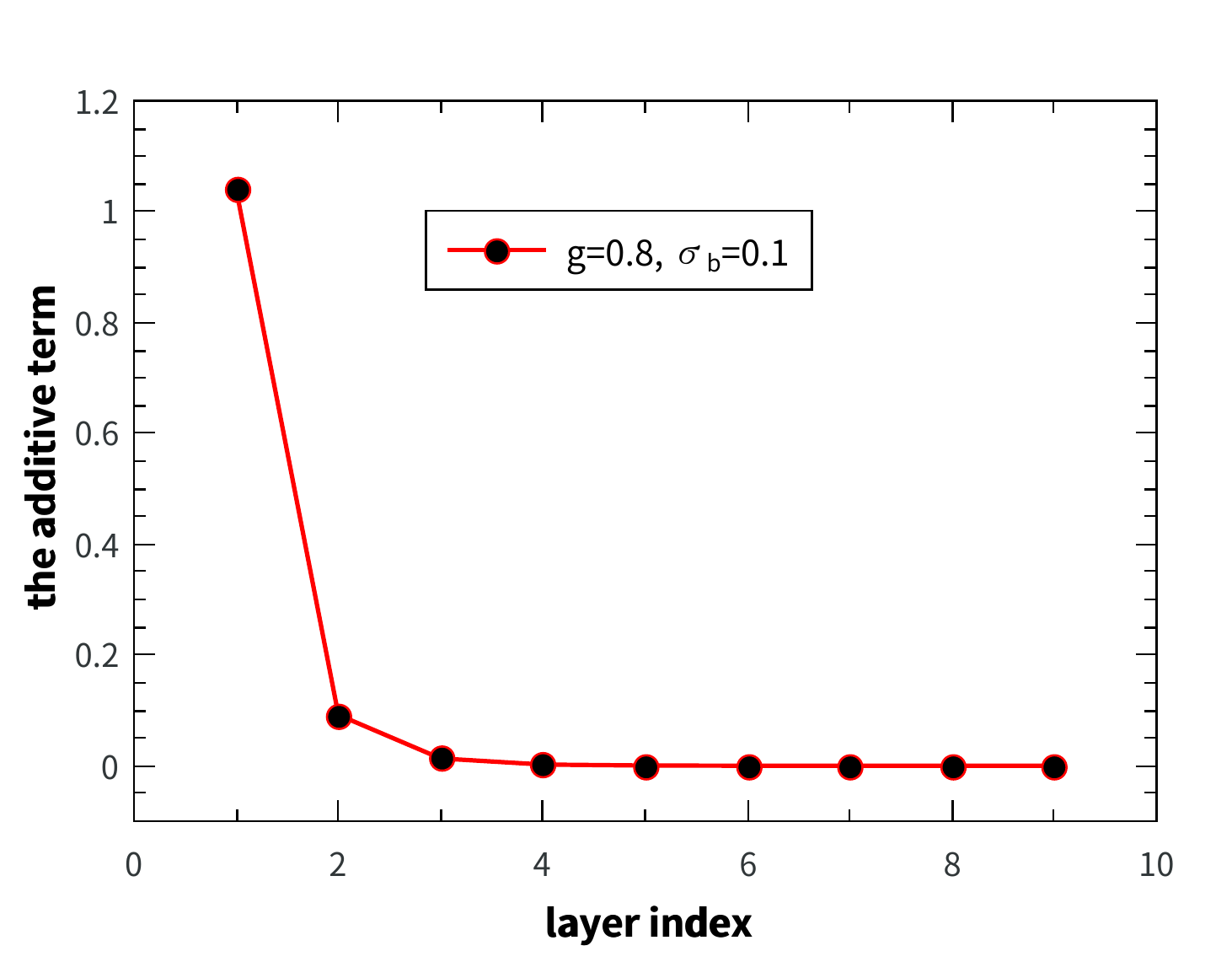}
  \caption{(Color online) The behavior of the additive term $(K_1^l)^2\Upsilon$ as a function of the network depth.
      }\label{additive}
 \end{figure}

To derive the relationship between dimensionality of consecutive layers, we first define $K_1^l=\frac{1}{N}\sum_iC_{ii}^l$ and $K_2^l=\frac{1}{N}\sum_i(C_{ii}^l)^2$,
then we get the normalized dimensionality of layer $l$ as
\begin{equation}\label{Dl}
 \tilde{D}^l=\frac{(K_1^l)^2}{(N-1)\Sigma^l+K_2^l},
\end{equation}
which is compared with the counterpart at a higher layer $l+1$ given by
\begin{equation}\label{Dlp1}
 \tilde{D}^{l+1}=\frac{(K_1^l)^2}{(N-1)\Sigma^l+K_2^l+(K_1^l)^2\Upsilon}.
\end{equation}
To derive Eq.~(\ref{Dlp1}), we used Eq.~(\ref{sigmal}).
Because the additive term $(K_1^l)^2\Upsilon$ in the denominator is always positive, Eq.~(\ref{Dlp1}) explains the dimensionality reduction across layers. The value of the additive
term thus determines how significantly the dimensionality is reduced. Its behavior with increasing number of layers can also be analyzed within the large-$N$ expansion.
First, we derive the recursion equation for $K_1^l$. Using the fact that $\Delta_{ii}\simeq gK_1^{l-1}$, one derives that
\begin{equation}\label{Kl}
 K_1^l=\int Du\int Dt\phi^2\Bigl(\sqrt{g(K_1^{l-1}+Q^{l-1})}t+\sqrt{\sigma_b}u\Bigr)-Q^l,
\end{equation}
by following the same principle as mentioned above.
Second, according to the definition, it is easy to write that 
\begin{equation}\label{psi}
 \Upsilon=\frac{\int Dt\int Du\Bigl(\phi'(\sqrt{\sigma_b}u+\sqrt{gQ^{l-1}}t)\Bigr)^4}{\Bigl(\int Dt\int Du(\phi'(\sqrt{\sigma_b}u+\sqrt{gQ^{l-1}}t))^2\Bigr)^2}.
\end{equation}
 Lastly, we find that
the additive positive term tends to be a very small value as the number of layers increases (Fig.~\ref{additive}), which is consistent with the observation in a finte-$N$ system. This implies that, the estimated
dimensionality at deep layers becomes nearly a constant, due to the nearly vanishing additive term.

\section{Training procedure of deep belief networks}
\label{train}
A deep belief network (DBN) is composed of multiple restricted Boltzmann machines (RBMs) stacked on top of each other. It is a probabilistic deep generative model, because after network parameters (weights and biases)
are learned (so-called training) from a data distributio, the model can be used to reproduce the samples mimicking the data distribution. With deep layers, the network becomes more expressive to capture 
high-order interdependence among components of a high-dimensional input, compared with a shallow RBM network.

Learning in a deep belief network can be achieved by layer-wise training of each RBM in a bottom-up pass, which was justified to
improve a variational lower bound on the data log-likelihood~\cite{Hinton-2006b}. RBM is a two-layered neural network, where there are no lateral connections within each layer, and the bottom (top) layer is also named the visible (hidden) layer.
Therefore, the RBM is described by the following energy function (also named Hamiltonian in physics):
\begin{equation}\label{energyRBM}
E(\bs,\bsg)=-\sum_{i,a}s_aw_{ia}\sigma_i-\sum_ab_a^hs_a-\sum_ib_i^v\sigma_i,
\end{equation}
where $\bs$ and $\bsg$ are the hidden and visible activity vector, respectively. $\mathbf{b}^{h,v}$ is the hidden (h) or visible (v) bias vector.
In statistical mechanics, the neural activity follows a Boltzmann distribution $P(\bs,\bsg)=\exp(-E(\bs,\bsg))/Z$, where $Z$ is the partition function of the model.
For a large network, $Z$ can only be computed by approximated methods~\cite{Huang-2015b}. This distribution can be used to fit any arbitrary discrete distribution, following 
the maximal likelihood learning principle, i.e., the network parameters are updated according to gradient ascent of the data log-likelihood defined as follows:
\begin{equation}\label{LLRBM}
\mathcal{L}=\sum_a\left<\ln\left(2\cosh\Bigl(\sum_iw_{ia}\sigma_i+b_a^h\Bigr)\right)\right>_{data}+\sum_ib_i^v\left<\sigma_i\right>_{data}-\ln Z,
\end{equation}
where the average is performed over all training data samples (or a mini-batch of the entire dataset in the case of stochastic gradient ascent used).
The gradient ascent leads to the following learning equations for updating network parameters:
\begin{subequations}\label{RBMLeq}
\begin{align}
\Delta w_{ia}&=\eta\left(\left<\sigma_is_ap(s_a|\bsg)\right>_{data}-\left<s_a\sigma_i\right>_{model}\right),\\
\Delta b_i^v&=\eta\left(\left<\sigma_i\right>_{data}-\left<\sigma_i\right>_{model}\right),\\
\Delta b_a^h&=\eta\left(\left<s_ap(s_a|\bsg)\right>_{data}-\left<s_a\right>_{model}\right),\\
\end{align}
\end{subequations}
where $\eta$ specifies a learning rate. Since there are no lateral connections between neurons at each layer, given one layer's activity, the other layer's activity
is factorized as 
\begin{equation}\label{factor}
p(\bs|\bsg)=\prod_ap(s_a|\bsg)=\prod_a\frac{e^{s_a([\bw\bsg]_a+b_a^h)}}{2\cosh([\bw\bsg]_a+b_a^h)}.
\end{equation}
Similarly, $p(\bsg|\bs)$ is also factorized. 

There are many approximate methods to evaluate the model-dependent terms in the 
learning equations. Here, we use the most popular method, namely contrast divergence~\cite{Hinton-2006b}. More precisely, RBMs are trained in a feedforward fashion 
using the contrast divergence algorithm~\cite{Hinton-2006b}, where Gibbs samplings of the model starting from each data point are truncated to a few steps, and then used to
compute model-dependent statistics for learning. The upper layer is trained with the lower layer's parameters being frozen. During the training of each RBM, the visible inputs are set to
the mean activity of hidden neurons at the lower layer, while hidden neurons of the upper layer still adopt stochastic binary values according to Eq.~(\ref{factor}). With this layer-wise training,
each layer learns a non-linear transformation of the data, and upper layers are conjectured to learn more abstract (complex) concepts, which is a key step in object and speech recognition problems~\cite{Lee-2011}.

The DBN learns a data distribution generated by a random RBM whose parameters follow the normal distribution $\mathcal{N}(0,g/N)$ for
weights and $\mathcal{N}(0,\sigma_b)$ for biases. $g=0.8$ and $\sigma_b=0.1$ unless otherwise specified. Using the RBM as a data generator allows us to
control the complexity of the input data. In addition, RBM has been used to model many real datasets (e.g., handwritten digits~\cite{Hinton-2006b}).
In numerical simulations (Fig. 3 in the main text), we generate $M=60000$ training examples (each example is an $N$-dimensional vector) from the random RBM. Then these examples are learned by RBMs in the DBN. 
We divide the entire dataset into mini-batches of size $B=150$.
One epoch corresponds to a sweep of the entire dataset.
Each RBM is trained for tens of epochs until the reconstruction error ($\varepsilon\equiv\|\bh'-\bh\|^2_2$) between input $\bh$ and reconstructed one $\bh'$ does not decrease.
We use an initial learning rate of $0.12$ divided by $\lceil t/10\rceil$ at $t$-th epoch, and an $\ell_2$ weight decay parameter of $0.0025$.

\section{Estimating the covariance structure of a random restricted Boltzmann machine}
\label{CMrbm}
To study the complexity propagation in the DBN, it is necessary to evaluate the statistics of the input data distribution, which is provided by a random RBM in the model setup of
stochastic neural networks. This is because, the estimated covariance can be used as a starting point from which the mean-field complexity-propagation equation iterates. 
In addition, characterizing the RBM representation may provide insights towards deep representations, 
since RBM is a building block for deep models and moreover a universal approximator of discrete distributions~\cite{Bengio-2008}.

Given the RBM, the hidden neural activity at a higher layer (e.g., $\bs$) can be marginalized over using the conditional independence (Eq.~(\ref{factor})), thus the distribution of the representation at a lower layer (e.g., $\bsg$) can 
be expressed as
\begin{equation}\label{rbm}
 P(\bsg)=\sum_{\bs}P(\bs,\bsg)=\frac{1}{Z}\prod_a\left[2\cosh([\bw^{l+1}\bsg]_a+b_a^{h})\right]\prod_ie^{\sigma_ib_i^v},
\end{equation}
where $Z$ is the partition function intractable for a large $N$. To study the statistics of the RBM representation, we need to compute the free energy function of Eq.~(\ref{rbm}) defined as $F=-\ln Z$, where a unit inverse temperature is assumed. We use the Bethe approximation to
compute an approximate free energy defined by $F_{{\rm bethe}}$. In physics, the Bethe approximation assumes $P(\bsg)\approx\prod_aP_a(\bsg_{\partial a})\prod_iP_i(\sigma_i)^{1-N}$~\cite{MM-2009}, where $a$ ($\partial a$) indicates a factor node (its neighbors)
representing
the contribution of one hidden neuron to the joint probability (Eq.~(\ref{rbm})) in a factor graph representation~\cite{Huang-2015b}. $P_a$ and $P_i$ can be obtained from a variational principle of free energy optimization~\cite{Huang-2015b}.
This approximation takes into account the correlations induced by nearest neighbors of each neuron in the factor graph,  which thus improves the naive mean-field approximation where neurons are assumed independent.

Covariance of neural activity under Eq.~(\ref{rbm}) (the so-called connected correlation in physics) can be computed from the approximate free energy using the linear response theory. However, due to the approximation, there exists a 
statistical inconsistency for diagonal terms computed under the Bethe approximation, i.e., $C_{ii}\neq1-m_i^2$. Therefore, we impose the statistical consistency of diagonal terms on
a corrected free energy as $\tilde{F}_{{\rm bethe}}=F_{{\rm bethe}}-\frac{1}{2}\sum_i\Lambda_i(1-m_i^2)$~\cite{atap-2013,Jack-2013,Yasuda-2013}. Following the similar procedure in
our previous work~\cite{Huang-2015b}, we obtain the following mean-field iterative equation:
\begin{subequations}\label{bprbm}
\begin{align}
m_{i\rightarrow a}&=\tanh\left(b_i^v-\Lambda_im_i+\sum_{a'\in\partial i\backslash
a}u_{a'\rightarrow i}\right),\\
u_{a'\rightarrow i}&=\frac{1}{2}\ln\frac{\cosh(b_{a'}^h+G_{a'\rightarrow
i}+w_{ia'})}{\cosh(b_{a'}^h+G_{a'\rightarrow i}-w_{ia'})},
\end{align}
\end{subequations}
where $G_{a'\rightarrow i}\equiv\sum_{j\in\partial a'\backslash i}w_{ja'}m_{j\rightarrow a'}$, and the correction introduces an Onsager term ($-\Lambda_im_i$). 
The cavity magnetization $m_{i\rightarrow a}$ can be
understood as the message passing from visible node $i$ to
factor node $a$, while the cavity bias $u_{a'\rightarrow i}$ is interpreted as the message
passing from factor node $a'$ to visible node $i$. In fact, Eq.~(\ref{bprbm}) is not closed. $\{\Lambda_i\}$ must be computed based on
correlations. Therefore, we define a cavity susceptibility $\chi_{i\rightarrow a,k}\equiv\frac{\partial m_{i\rightarrow a}}{\partial b_k^v}$~\cite{susp-2010}.
According to this definition and the linear response theory, we close Eq.~(\ref{bprbm}) by obtaining the following susceptibility propagation equations:
\begin{subequations}\label{sprbm}
\begin{align}
\begin{split}
\chi_{i\rightarrow a,k}&=(1-m^2_{i\rightarrow a})\sum_{a'\in\partial i\backslash a}\Gamma_{a'\rightarrow i}\mathcal{P}_{a'\rightarrow i,k}\\
&+\delta_{ik}(1-m^2_{i\rightarrow a})-
\Lambda_iC_{ik},
\end{split}\\
C_{ik}&=\frac{1-m_i^2}{1+(1-m_i^2)\Lambda_i}\mathcal{F}_{ik},\\
\Lambda_i&=\frac{\mathcal{F}_{ii}-1}{1-m_i^2},
\end{align}
\end{subequations}
where the full magnetization $m_{i}=\tanh\left(b_i^v-\Lambda_im_i+\sum_{a'\in\partial i}u_{a'\rightarrow i}\right)$, 
$\Gamma_{a\rightarrow i}\equiv\frac{\tanh(w_{ia})(1-\tanh^2(b_{a}^h+G_{a\rightarrow
i}))}{1-\tanh^2(b_{a}^h+G_{a\rightarrow
i})\tanh^2(w_{ia})}$, $\mathcal{P}_{a\rightarrow i,k}\equiv\sum_{j\in\partial a\backslash i}\chi_{j\rightarrow a,k}w_{ja}$, and $\mathcal{F}_{ik}\equiv
\sum_{a\in\partial i}\Gamma_{a\rightarrow i}\mathcal{P}_{a\rightarrow i,k}+\delta_{ik}$. It is easy to verify that Eq.~(\ref{sprbm}) leads to the consistency for the 
diagonal terms. Adding the diagonal constraint through Lagrange multiplier $\bm{\Lambda}$ can not only solve the diagonal inconsistency problem but also improve the 
accuracy of estimating off-diagonal terms. After the RBM parameters (weights and biases) are specified, we run the above iterative equations (Eq.~(\ref{bprbm}) and Eq.~(\ref{sprbm})) from a random
initialization of the messages, and estimate the covariance and associated representation dimensionality from the fixed point. These statistics are used as 
an initialization condition for the complexity propagation equation (Eq.(1) in the main text) that is used to study the expressive property of the DBN with trained weights and biases. 

We finally remark that, for a \textit{trained} RBM, some components of the correlation matrix may
lose the symmetry property ($C_{ij}\neq C_{ji}$), likely because of the above Bethe (cavity-based) approximation incapable of dealing with an irregular distribution of learned connection weights. The irregularity means that
the distribution is divided into two parts: the bulk part is around zero, while the other part is dominated by a few large values of weights (as also observed recently in spectral dynamics of learning in RBM~\cite{Decelle-2017}). 
Our mean-field formula (Eq.~(\ref{bprbm}) and Eq.~(\ref{sprbm})) may offer a basis to be
further improved to address this interesting special property, although one can enforce the symmetry by $[C_{ij}+C_{ji}]/2$ in our mean-field formula.

\begin{figure}
  \includegraphics[bb=0 0 351 244,scale=0.7]{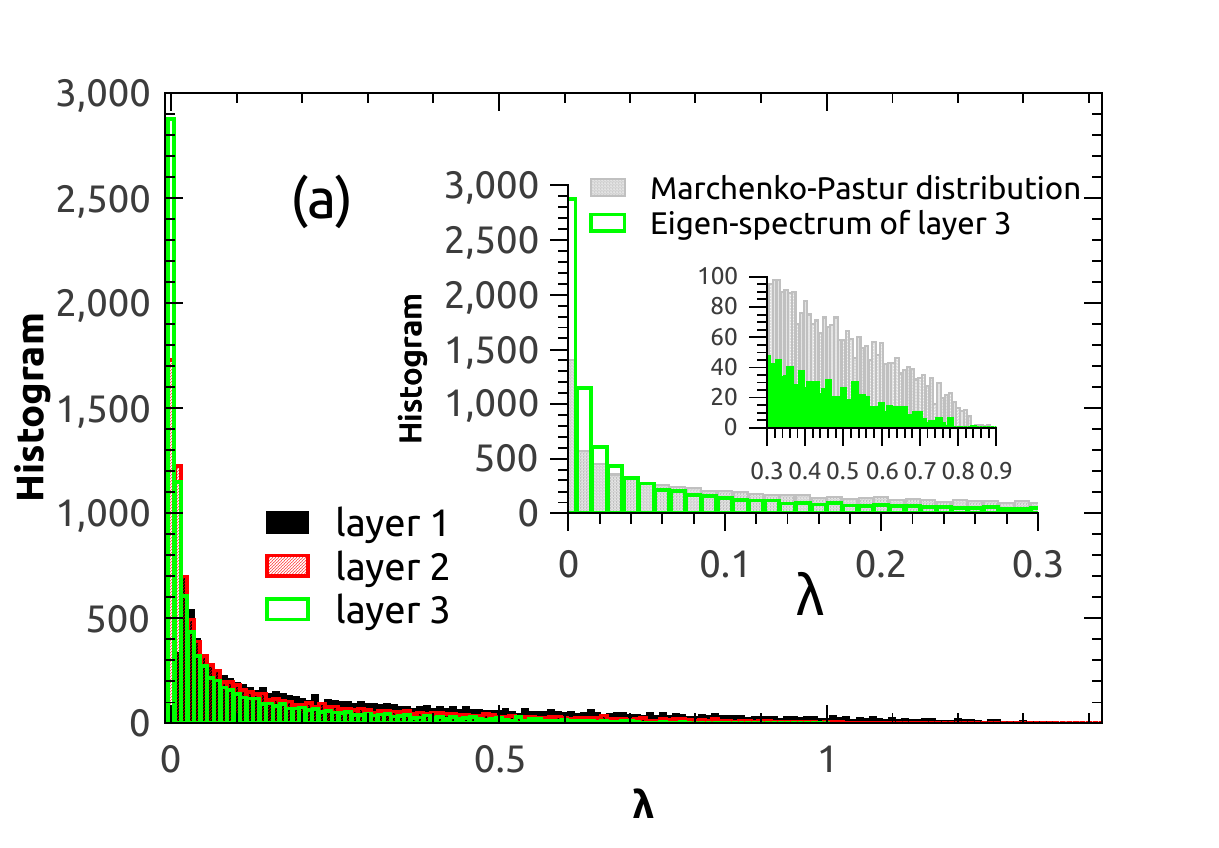}
  \includegraphics[bb=0 0 339 244,scale=0.7]{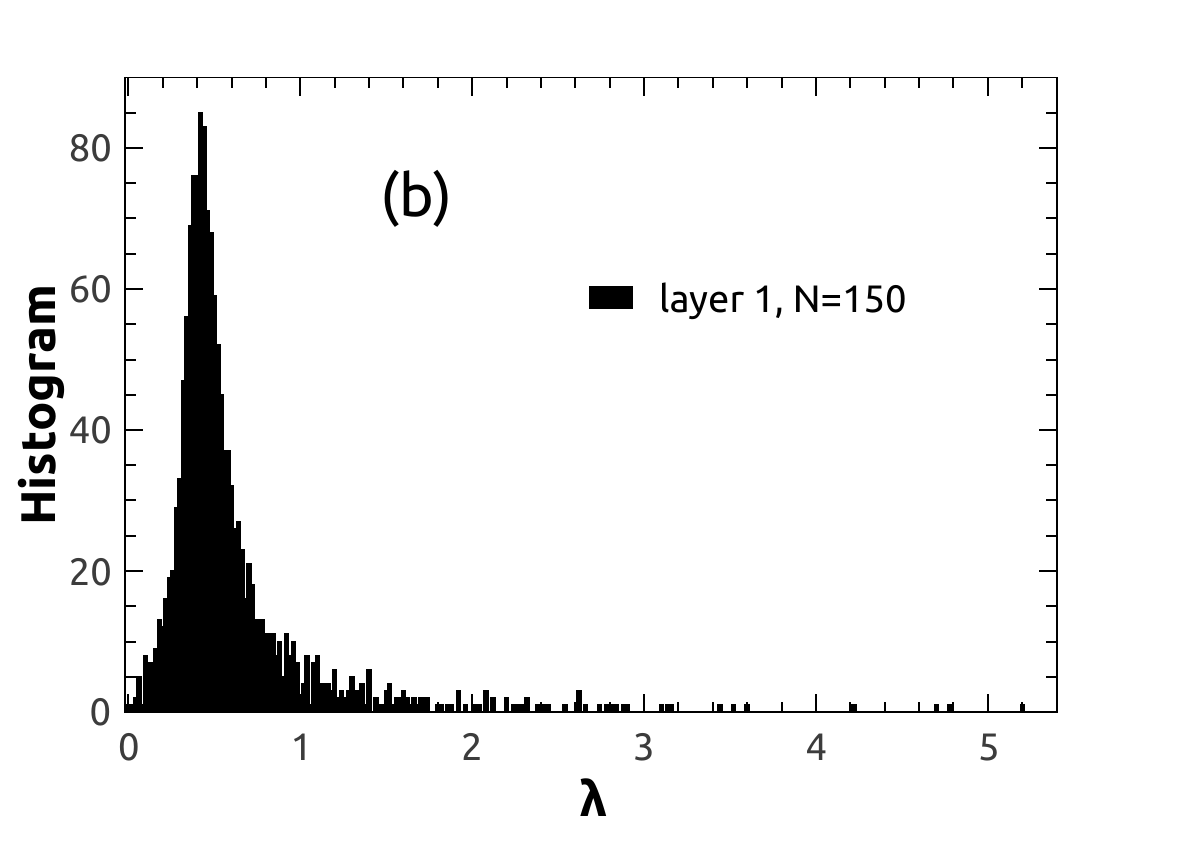}
  \caption{(Color online) The eigenvalue distribution of the covariance matrix estimated from the deep computation. (a) The distribution for the deep networks with random weights ($N=100$). One hundred instances are used. In the inset, the distribution at deeper layers is compared with the Marchenko-Pastur law of a random-sample 
  covariance matrix. The tail part is enlarged for comparison. (b) The distribution obtained from an unsupervised learning system of deep belief networks is strongly different from the Marchenko-Pastur law. Ten instances are used.
      }\label{spectral}
 \end{figure}

\section{The eigenvalue distribution of the covariance matrix}
\label{spect}
To analyze the eigenvalue distribution of the covariance matrix at each layer of deep networks, we first construct a random-sample covariance matrix. More precisely,
we consider a real Wishart ensemble, where a random-sample covariance matrix is defined as $\frac{1}{N}\boldsymbol{\xi}\boldsymbol{\xi}^{{\rm T}}$ in which $\boldsymbol{\xi}$ defines 
an $N\times P$ matrix whose entries follow independently a normal distribution $\mathcal{N}(0,\varsigma^2)$. In fact, $\boldsymbol{\xi}$ can be thought of as a random uncorrelated pattern matrix. To
compare the real Wishart ensemble with the covariance matrix estimated from the mean-field theory of deep networks, we choose $P=N$, and $\varsigma^2$ is obtained by matching the range of the eigenvalue.
The designed random-sample covariance matrix has the Marchenko-Pastur law for the density of eigenvalues~\cite{Dean-2003,MPlaw}:
\begin{equation}
 \mu(\lambda)=\frac{1}{2\pi\lambda\varsigma^2}\left[(\lambda-\lambda_{-})(\lambda_{+}-\lambda)\right]^{1/2},
\end{equation}
where $\lambda\in[\lambda_{-},\lambda_{+}]$, and $\lambda_{-}=0$, $\lambda_{+}=4\varsigma^2$.

For the random neural networks, the eigenvalue distribution at deep layers seems to assign a higher probability density when the eigenvalue gets close to zero, yet
a lower density at the tail of the distribution (Fig.~\ref{spectral} (a)), compared with the Marchenko-Pastur law of a random-sample covariance matrix. For the unsupervised learning system, we find that the distribution deviates significantly from
the Marchenko-Pastur law of a Wishart ensemble. The distribution has a Gaussian-like bulk part together with a long tail (Fig.~\ref{spectral} (b)). Therefore, the dimensionality reduction and its relationship with decorrelation
are a nontrivial result of the deep computation.



\end{document}